\documentclass[runningheads]{llncs}

\usepackage{graphicx}
\usepackage{xcolor}
\usepackage{subcaption}
\usepackage{hyperref}
\usepackage{verbatim}

\makeatletter
\newcommand{\printfnsymbol}[1]{%
  \textsuperscript{\@fnsymbol{#1}}%
}
\makeatother

\begin{document}

\title{\textit{Physio}: An LLM-Based Physiotherapy Advisor}

\author{
    Rúben Almeida\thanks{Equal contribution.} \inst{1} \orcidID{0000-0002-1942-2399} \and
    Hugo Sousa\printfnsymbol{1} \inst{1,2} \orcidID{0000-0003-3226-9189} \and 
    Luís F. Cunha\printfnsymbol{1} \inst{1,2} \orcidID{0000-0003-1365-0080} \and
    Nuno Guimarães\inst{1,2} \orcidID{0000-0003-2854-2891} \and 
    Ricardo Campos \inst{1,3,4} \orcidID{0000-0002-8767-8126} \and 
    Alípio Jorge \inst{1,2} \orcidID{0000-0002-5475-1382}
}

\authorrunning{R. Almeida, H. Sousa, L. Cunha et al.}

\institute{
    INESC TEC, Porto, Portugal \and
    University of Porto, Porto, Portugal \and
    University of Beira Interior, Covilhã, Portugal \and 
    Ci2 - Smart Cities Research Centre, Tomar, Portugal 
    \email{\{ruben.f.almeida,luis.f.cunha,hugo.o.sousa, \\ nuno.r.guimaraes,ricardo.campos,alipio.jorge\}@inesctec.pt}
}

\maketitle

\begin{abstract}
    The capabilities of the most recent language models have increased the interest in integrating them into real-world applications. However, the fact that these models generate plausible, yet incorrect text poses a constraint when considering their use in several domains. Healthcare is a prime example of a domain where text-generative trustworthiness is a hard requirement to safeguard patient well-being. In this paper, we present Physio, a chat-based application for physical rehabilitation. Physio is capable of making an initial diagnosis while citing reliable health sources to support the information provided. Furthermore, drawing upon external knowledge databases, Physio can recommend rehabilitation exercises and over-the-counter medication for symptom relief. By combining these features, Physio can leverage the power of generative models for language processing while also conditioning its response on dependable and verifiable sources. A live demo of Physio is available at \url{https://physio.inesctec.pt}.

\end{abstract}

\section{Introduction}
Although language models (LMs) have long been studied by the research community, they only reached mainstream attention with the release of the ChatGPT application by OpenAI~\cite{ChatGPT}. This application granted the public access to a highly effective generative model, GPT-3.5~\cite{ouyang2022training}, that was capable of producing coherent conversations on various topics, a novelty at the time. This development naturally led to the emergence of numerous applications and discussions regarding the potential applications of generative models in various domains, such as law~\cite{savelka_unlocking_2023}, education~\cite{savelka_can_2023}, and health~\cite{kung_performance_2023,levine_diagnostic_nodate}. However, these models also exhibited significant limitations that hindered their implementation in those domains. At the top of that list is the hallucination problem~\cite{openai2023gpt4}, \textit{i.e.}, their propensity to generate incorrect yet convincing answers. This limitation prompted increased research into grounding the text generated by these models on reliable sources, a task known as retrieval-augmented generation~\cite{NEURIPS2020_6b493230}. The general approach starts by retrieving documents relevant to the input query and subsequently using them to generate an answer. By doing so, one can link the generated texts to the original documents, thereby providing the user references where he/she can get more information supporting the generated answer~\cite{gao2023enabling,huang2023citation}. This research gave rise to systems like BingChat~\cite{BingChat} and Bard~\cite{Bard}, search engines that combine the personalization of answers generated by LMs with the trustworthiness provided by the retrieval component of the system. This concept can be taken one step further to be applied to domain-specific applications by constraining the retrieval component to a specialized set of documents. This is the main idea behind the demo we present in this paper, Physio, a chat-based application tailored to help users in physical rehabilitation. Physio's answers are generated from an augmented prompt that compiles the user input with documents from a curated knowledge base specifically tailored to only contain reliable sources for physical rehabilitation. Furthermore, the sentences in the generated answer present references to those documents. Apart from that, Physio's answer can also contain exercise and over-the-counter medication recommendations whenever appropriate. The source code for Physio is open-source and available on GitHub\footnote{\url{https://github.com/liaad/physio}}.

\section{Physio}\label{sec:physio}
\begin{figure}[ht]
    \centering
    \includegraphics[width=\textwidth]{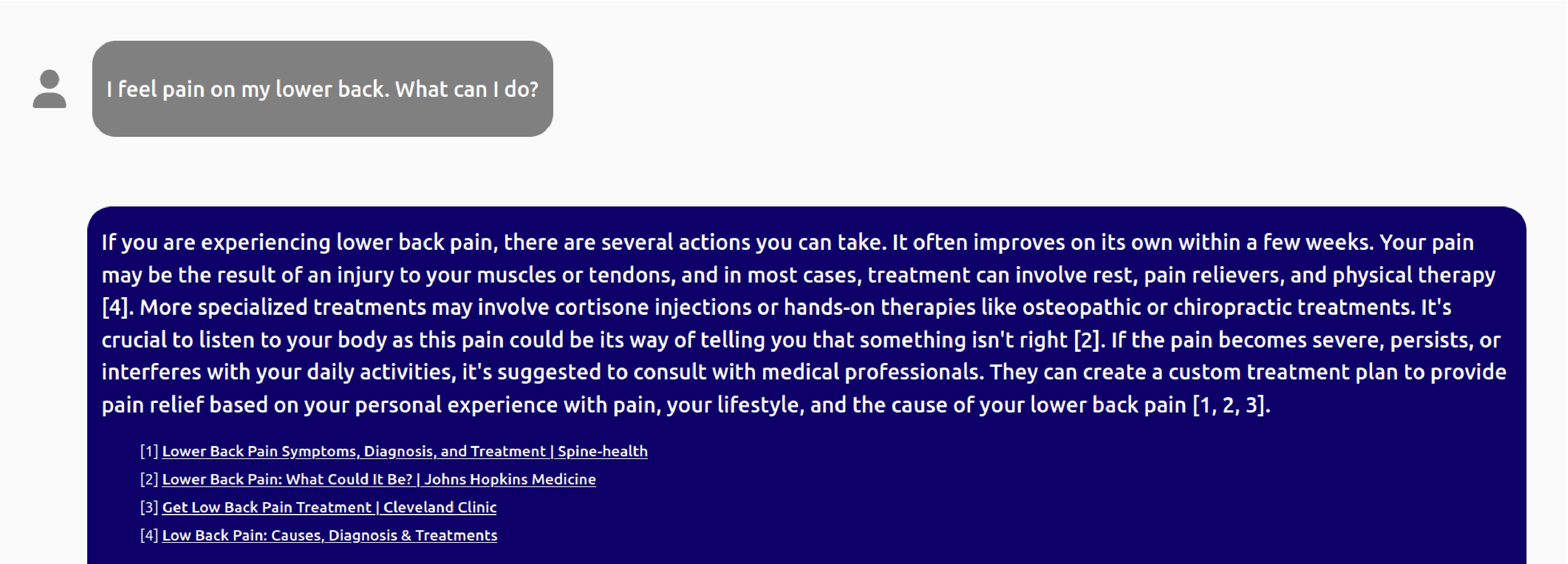}
    \caption{Screenshot from Physio web demonstration. The user input is in the grey box, while the system answer is presented in the blue box.}
    \label{fig:answer}
\end{figure}

To some extent, Physio can be thought of as an initial step toward an artificial intelligent physiatrist. To this end, we designed it to offer functionalities akin to those found in a physiatrist appointment, specifically: explaining/discussing the problem that the user is querying; and recommending exercises and medication when deemed appropriate. For the LM, Physio relies on OpenAI GPT-4 model which is the most effective LM at the time of development of this demo~\cite{openai2023gpt4}. An example answer from Physio is illustrated in Figure \ref{fig:answer} for the prompt ``I feel pain in my lower back. What can I do?'' In the following sections, we explain how Physio's answer is built. 

\subsection{Knowledge-base Construction} 
The construction of the knowledge for Physio commenced by scraping the Rehab Hero website~\cite{RehabHero}. This website boasts an extensive compilation of exercises, each accompanied by an instructional video demonstrating execution, and information on the specific physical conditions they address. After scraping this website, we compiled all the physical conditions that we found on the Rehab Hero pages and queried each of them in a set of $20$ websites that provided more information about the condition\footnote{More details about the list of websites selected can be found on the GitHub repository.}. To ensure the reliability of our system the list of websites was curated and validated by a physiotherapist and includes sources like Mayo Clinic~\cite{MayoClinic}, NHS~\cite{NHS}, and OrthoInfo~\cite{OrthoInfo}. For the medication-related aspects, we rely on the DrugBank database~\cite{wishart_drugbank_2018}. The information obtained was then indexed in a MongoDB database composed of three collections: \texttt{exercises}, \texttt{webpages}, and \texttt{medications}. The first two are indexed based on the associated condition, while the last is indexed by drug name.

\subsection{Data Pipeline} When a user submits a query in Physio, the text undergoes processing through a data pipeline to generate a response. The initial step in this pipeline is to \textbf{verify if it is an English physiotherapy-related prompt}. This validation is achieved by using the LM and a predefined validation prompt template that assesses whether the user's input is related to physiotherapy and written in English. The validation prompt template conditions the LM's response to produce a boolean output (either ``True'' or ``False'') so that it can be programmatically interpreted. If the input is deemed invalid, the system provides a default response. Instead, if the input is validated, the application proceeds to the \textbf{condition identification} step, where it determines the condition in the user's question. This is accomplished by employing a few-shot template to instruct the LM to identify the condition. For instance, when presented with the input ``I have sprained my ankle'' the model should identify the condition as ``ankle sprain.'' 

Once the condition is identified, it is \textbf{linked to one of the entries in our database}. This linkage process first attempts an exact match, followed by a search in the list of aliases (for instance ``lumbago'' is in the list of aliases for ``back pain''), and, as a last resort, employs substring matching. In case no match is found, the LM is prompted with the user query, and the answer is returned. Otherwise, the pipeline advances to the generation of the answer and extraction exercises and medication.

\paragraph{\normalfont \textbf{Answer Generation}} With the linked condition one can retrieve the list of documents related to that condition from the \texttt{webpages} collection. Among the pages available, we employ the BM25 retrieval model~\cite{Amati2009} to search and rank them based on their relevance to the user's input. The top five ranked documents are subsequently provided to the generative model, along with the user's input query, using a prompt template designed to instruct the model to answer user questions using the information contained in these pages.

To provide the user a way to verify the trustworthiness of the generated text, a list of references is incorporated by ranking the sentences from the original source pages in relation to the sentences in the generated text, again, using the BM25 ranking method. Note that determining the optimal number of references to include in the final answer is not trivial. In our application, we establish a heuristic to use as reference the top-N ranked sentence-document pairs, where N is the number of generated sentences.\footnote{While this heuristic has been effective in practice, ongoing research is aimed at refining the reference selection process based on similarity scores.} However, the final answer may not necessarily contain the same number of references as sentences, as a generated sentence can be highly similar to multiple sentences within a given document.

\paragraph{\normalfont \textbf{Exercise \& Medication Extraction}} The linked condition is also used to fetch the exercises directly from the \texttt{exercises} collection, as they are indexed by condition. For the linked condition we randomly sample up to five exercises to be presented in the web interface.

The final element of the response pertains to medication recommendations. This is accomplished by instructing the LM to provide medication suggestions based on the user's query, the linked condition, and the generated answer. The prompt explicitly specifies that the response should be in the form of a JSON-parsable list of strings, where each string represents a medication. After parsing these strings, we conduct a search for the recommended drugs within the \texttt{medication} collection of our database, first by exact matching and subsequently with fuzzy matching. \\

The last task of the pipeline is to combine the three components of the answer, send them to the frontend, and cache the result in the database. 

\paragraph{\normalfont \textbf{Ethical Considerations}} Given the sensitive nature of this domain, ethical considerations are paramount. As a result, we include a disclaimer on Physio's website, explicitly stating that it is a research demonstration, and we strongly advise users to consult with a specialist before making any decisions regarding their health. Furthermore, we have limited medication recommendations to include only over-the-counter options.

{\footnotesize \paragraph{\normalfont \textbf{Aknowledgments}} This work is financed by National Funds through the Fundação para a Ciência e a Tecnologia, within the project StorySense (DOI \url{10.54499/2022.09312.PTDC}) and the Recovery and Resilience Plan within the scope of the Health From Portugal project.}

\bibliographystyle{splncs04}
\bibliography{references}

\end{document}